\documentclass{article}
\usepackage{iclr2027_conference,times}

\usepackage{amsmath,amsfonts,bm}

\def\eqref#1{equation~\ref{#1}}

\def\1{\bm{1}}

\DeclareMathAlphabet{\mathsfit}{\encodingdefault}{\sfdefault}{m}{sl}
\SetMathAlphabet{\mathsfit}{bold}{\encodingdefault}{\sfdefault}{bx}{n}

\usepackage{hyperref}
\usepackage{url}
\usepackage{graphicx}
\usepackage{algorithm}
\usepackage{algorithmic}
\usepackage{threeparttable}
\usepackage{booktabs}
\usepackage{amssymb}
\usepackage{multirow}
\usepackage{adjustbox}
\usepackage{enumitem}
\usepackage{pifont}
\usepackage{wrapfig}
\usepackage{capt-of}
\usepackage{comment}
\usepackage{tabularx}

\usepackage[table]{xcolor}

\usepackage{xcolor}
\usepackage{listings}

\definecolor{promptbg}{RGB}{246,246,246}

\lstdefinestyle{affectprompt}{
  basicstyle=\ttfamily\footnotesize,
  backgroundcolor=\color{promptbg},
  frame=none,
  breaklines=true,
  breakatwhitespace=false,
  columns=fullflexible,
  keepspaces=true,
  showstringspaces=false,
  xleftmargin=0.7em,
  xrightmargin=0.7em,
  aboveskip=0.45em,
  belowskip=0.75em,
  literate={—}{{\textemdash}}1
}

\title{AffectReveal: Event-Grounded Emotion Recognition Beyond Visual Appearances}

\author{
Yihao Qian\textsuperscript{1},
Runhao Zeng\textsuperscript{2},
Sicheng Zhao\textsuperscript{3},
Feng Liang\textsuperscript{2},
Hongmin Cai\textsuperscript{1},
Mingkui Tan\textsuperscript{1},
\\[2pt]
\normalfont
\textsuperscript{1}South China University of Technology, Guangzhou, China\\
\textsuperscript{2}Shenzhen MSU-BIT University, Shenzhen, China\\
\textsuperscript{3}Tsinghua University, Beijing, China\\[2pt]
\texttt{202421045609@mail.scut.edu.cn}\\
}

\iclrfinalcopy
\begin{document}

\maketitle
\lhead{}

\begin{abstract}
Visual emotion recognition commonly assumes that all evidence required for
prediction is contained in the observed image or video. Yet the same visible
reaction can convey different emotions depending on events beyond the input:
tears, for example, may indicate grief or joy. We formulate
\textbf{Event-Grounded Emotion Recognition (EGER)}, where emotion recognition
requires recovering the affect-determining event. We construct
\textbf{EGER-Bench}, comprising 10,052 videos and 10,734 images across 11
emotions, two source domains, and four visual settings. A study with six
annotators shows that event context raises human recognition accuracy from
33.96\% to 72.08\%, confirming that visual evidence alone is often
insufficient. Semantic relevance alone does not solve EGER: a plausible event may imply the
wrong emotion if its identity, focal-person role, relationship, or outcome is
misinterpreted. We therefore propose \textbf{AffectReveal}, a tuning-free
framework that first constructs and independently verifies evidence-grounded
alternatives over these affect-critical factors. It then cross-checks the
recovered event against face-masked in-media facts through bidirectional
atomic evidence support, while retaining the original unmasked input for
final prediction. Across three downstream models and four input settings,
AffectReveal yields average UAR gains of 5.26--10.53 points. For three
fine-tunable models, it also enables untuned models to outperform their
fine-tuned visual-only counterparts in all 12 accuracy comparisons, without
updating downstream parameters.
\end{abstract}

\section{Introduction}
\label{sec:intro}

Visual emotion recognition infers a person's emotional state from facial expressions, body movements, interactions, and scene context~\citep{mollahosseini2017affectnet,kosti2017emotic,lee2019context}.
Despite substantial progress, it is still predominantly formulated as \emph{closed-input recognition}, assuming that the observed image or video contains all evidence required for emotion prediction.

\begin{figure*}[t]
    \centering
    \includegraphics[width=\textwidth]{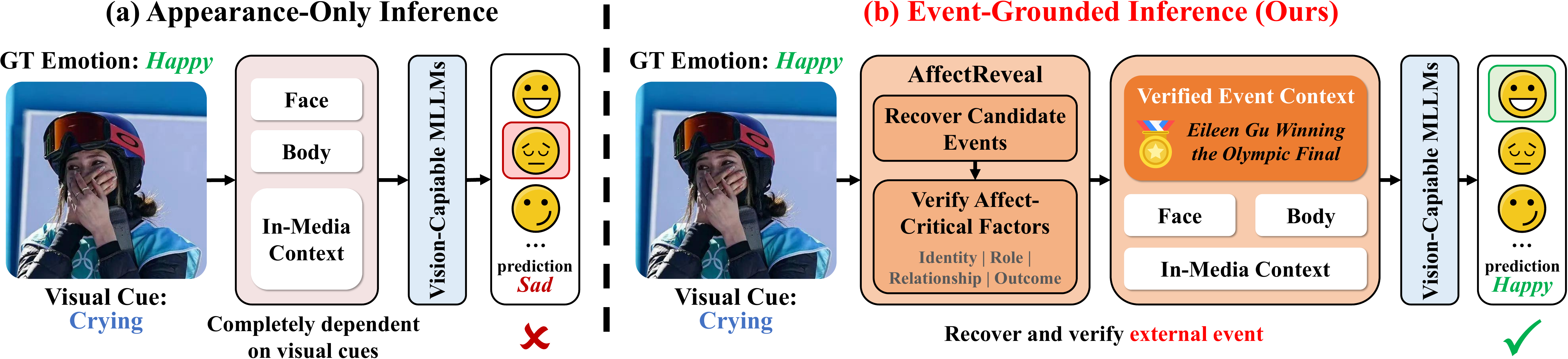}
    \vspace{-0.7cm}
    \caption{
    Appearance-based versus event-grounded emotion recognition.
    The visible reaction suggests sadness, whereas the recovered event context supports happiness.
    AffectReveal recovers and verifies the affect-determining event before prediction.
    }
    \label{fig:motivation}
\end{figure*}

This assumption does not always hold.
The same tears may arise from grief, relief, or victory, while the same smile may express happiness, embarrassment, or concealed distress.
A visible reaction can therefore imply different emotions depending on what happened and how the focal person was involved~\citep{xia2019emotion,poria2021recognizing}.
As illustrated in Figure~\ref{fig:motivation}, a crying athlete may appear sad from visual appearance alone, yet becomes understandable as happy once the underlying event---winning an Olympic final---is recovered.

Context-aware methods incorporate scenes, objects, actions, and social interactions~\citep{kosti2017emotic,kosti2019context,lee2019context,mittal2020emoticon}, but still reason over evidence contained in the given media.
They cannot resolve cases in which decisive information lies outside the observed sample, such as event identity, focal-person role, interpersonal relationship, or event outcome.
We study this evidence gap through \textbf{Event-Grounded Emotion Recognition (EGER)}, where recognizing an emotion requires recovering the external event needed to interpret the observed reaction.

To enable systematic evaluation, we construct \textbf{EGER-Bench}, comprising 10,052 video clips and 10,734 images across 11 emotions, two source domains, and four visual input settings.
The benchmark focuses on reactions whose short-form visual evidence is insufficient or potentially misleading.
A controlled human study confirms this information gap: providing event context increases emotion-recognition accuracy from 33.96\% to 72.08\%.

External retrieval is a natural starting point~\citep{lewis2020retrieval,yao2022react,asai2024self}, but introduces two task-specific challenges.
First, a retrieved event may be topically plausible yet imply the wrong emotion if its identity, focal-person role, relationship, or outcome is misinterpreted, which we call \emph{affect-critical event ambiguity}.
Second, using a potentially misleading expression to validate the event retrieved from that expression creates \emph{appearance-induced circularity}.

We propose \textbf{AffectReveal}, a tuning-free framework designed around these challenges.
\textbf{Affect-Critical Event Recovery} constructs and independently verifies evidence-anchored alternatives over the four affect-critical factors before consolidating the external event context.
\textbf{Expression-Disentangled Event Verification} then masks facial regions only in the verification branch and compares non-facial in-media facts with the recovered event through bidirectional atomic evidence support.
The original unmasked input and the verified event evidence are jointly used for final prediction.

Across three downstream models and four input settings, AffectReveal improves both UAR and accuracy in all 16 direct comparisons, with average UAR gains of 5.26--10.53 points.
An untuned model equipped with AffectReveal also outperforms its fine-tuned visual-only counterpart in all 12 corresponding accuracy comparisons.
Component analysis further shows that non-facial context becomes useful when employed to verify the recovered event, rather than simply appended as additional text.

Our contributions are:

\textbf{(1)} We formulate \textbf{Event-Grounded Emotion Recognition (EGER)}, where the affect-determining event is not fully contained in the observed media and must be recovered to interpret the focal person's visual reaction;

\textbf{(2)} We construct \textbf{EGER-Bench}, which pairs visually insufficient or misleading short-form observations with source-grounded event annotations across two domains and four controlled input settings, enabling systematic evaluation of emotion recognition under missing event evidence;

\textbf{(3)} We propose \textbf{AffectReveal}, a tuning-free framework that verifies emotion-relevant event details and cross-checks the recovered event with non-facial visual evidence, improving multiple downstream models without parameter updates.

\section{Related Work}
\label{sec:related_work}

\paragraph{Visual and Context-Aware Emotion Recognition.}
Visual emotion recognition has expanded from face-centered classification
~\citep{barsoum2016training,li2017reliable,jiang2020dfew,liu2022mafw} to
models incorporating body posture, scenes, objects, actions, and social
interactions~\citep{kosti2017emotic,kosti2019context,lee2019context,
hoang2021context,wu2022hierarchical}. Subsequent approaches further reduce
dependence on facial appearance through masking, context deconfounding, and
interaction modeling~\citep{mittal2020emoticon,yang2022emotion,
yang2023context,li2026towards}. However, their contextual evidence remains
contained in the observed media. EGER instead addresses reactions whose
affect-determining events are not fully observable in the input. Recent MLLMs
and emotion-oriented variants provide strong downstream reasoners, but the EGER challenge
concerns missing evidence rather than a particular model architecture~\citep{liu2023visual,bai2023qwen,chen2024internvl,
cheng2024emotion,lian2025affectgpt,peng2026emotion}.

\paragraph{Emotion Causes and Event-Level Affect Understanding.}
Emotion-cause research relates affective states to the situations that elicit
them~\citep{gui2016event,xia2019emotion,rashkin2019towards,zadeh2019social}.
Existing methods typically identify causes from a provided conversation,
document, or complete multimodal sequence
~\citep{poria2021recognizing,wang2022multimodal,li2025multimodal}. In EGER,
the relevant event is not given and must be recovered externally. The model
must therefore establish not only which event occurred, but also which role,
relationship, and outcome apply to the focal person.

\paragraph{Retrieval-Augmented Affect Reasoning.}
Retrieval-augmented and tool-using models acquire external evidence for
knowledge-intensive reasoning~\citep{lewis2020retrieval,guu2020retrieval,
izacard2021leveraging,yao2022react,schick2023toolformer,qin2024toolllm},
while iterative retrieval and critique improve factual support
~\citep{madaan2023self,asai2024self,gou2024critic}. Retrieval-augmented
multi-agent reasoning has also been explored for multimodal emotion
recognition~\citep{wang2026affectagent}. AffectReveal targets a distinct
affective failure mode: retrieved information can be topically relevant yet
imply the wrong emotion. It therefore verifies evidence-grounded alternatives
over affect-critical event details and cross-checks the recovered event
against non-facial visual evidence.

\section{Event-Grounded Emotion Recognition and EGER-Bench}
\label{sec:benchmark}

\subsection{Task Definition}
\label{sec:task_definition}

Let $x^{m}$ denote an image or video under visual setting
$m\in\mathcal{M}$, and let $y\in\mathcal{Y}$ be the emotion of a designated
focal person. EGER considers cases in which the emotion depends not only on
the observed reaction but also on a latent affect-determining event $z^{*}$:
\begin{equation}
    y=f(x^{m},z^{*}),\qquad
    \hat{z}=\mathcal{R}(x^{m};\mathcal{K}),\qquad
    \hat{y}=F(x^{m},\hat{z})\in\mathcal{Y}.
\label{eq:eger_task}
\end{equation}
At inference time, $z^{*}$ is unavailable. A method may instead query an
external information space $\mathcal{K}$ to recover an event interpretation
$\hat{z}$. Importantly, $\mathcal{K}$ is not sample-specific reference
context supplied with the test input. Emotion prediction is the primary task;
event recovery is evaluated separately through diagnostic analyses.

\begin{figure*}[t]
    \centering
    \includegraphics[width=\textwidth]{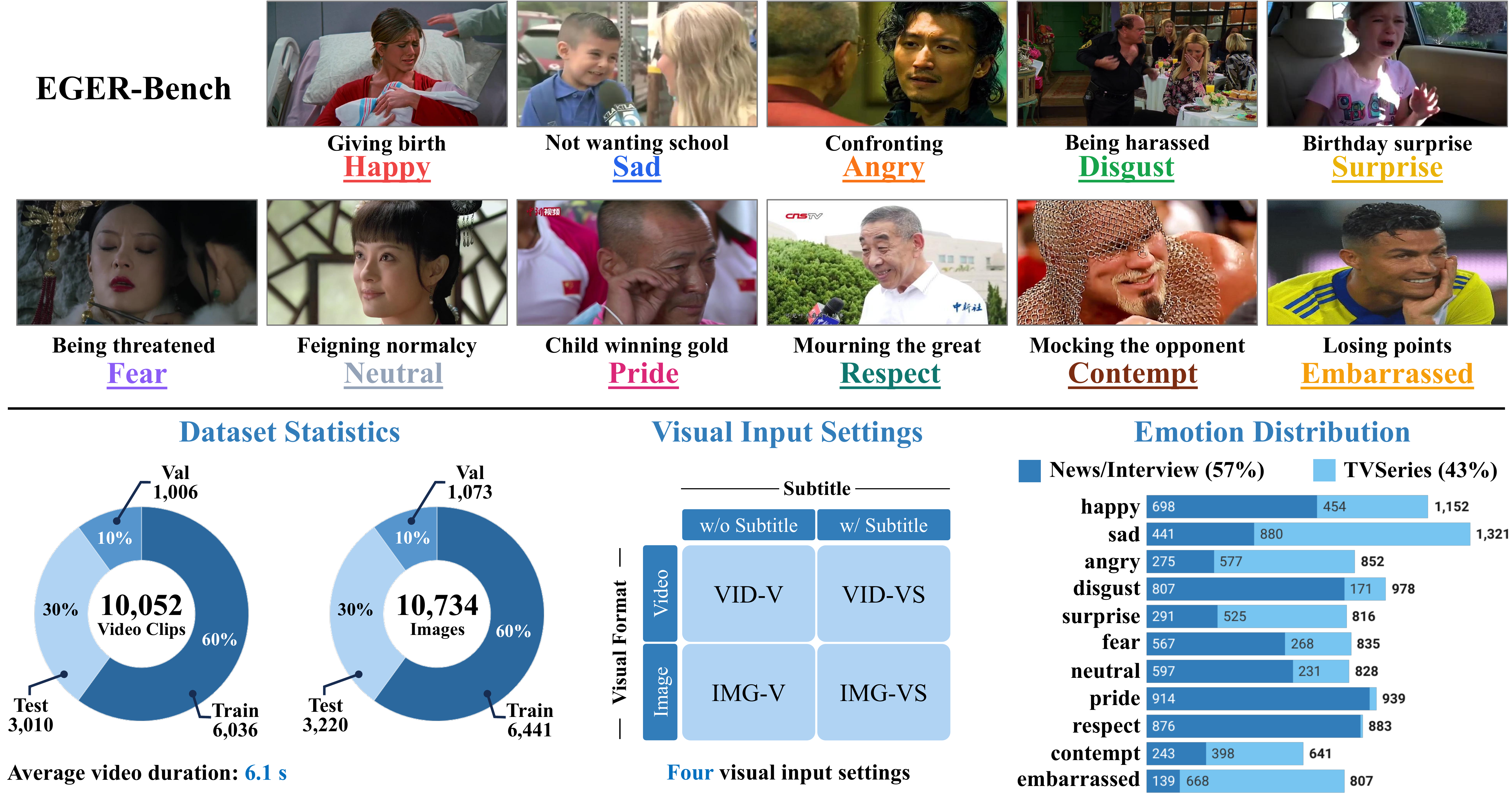}
    \vspace{-0.7cm}
    \caption{
    Overview of EGER-Bench, covering 11 emotions from news/interview and
    television-series sources. The lower panel summarizes dataset statistics,
    four visual settings, and class distributions across the two domains.
    }
    \label{fig:dataset}
\end{figure*}

\subsection{Data Collection}
\label{sec:data_collection}

EGER-Bench draws from two complementary domains. \textbf{News and interview}
sources provide real-world events documented by webpages and headlines,
whereas \textbf{television series}, including \textit{Friends} and
\textit{Empresses in the Palace}, provide richer relationships, intentions,
and role-dependent reactions.

We use seed patterns such as ``tears of joy,'' ``putting on a brave face,''
and ``forcing a smile'' to discover reactions whose emotional meaning may
differ from their appearance. For news and interview sources, we retrieve
candidate videos and extract coherent, person-centered clips shorter than
10 seconds. For television series, we identify candidates from dialogue
scripts, align them with episodes using subtitles, and extract clips centered
on the focal person. These patterns are used only for discovery; inclusion
and annotation are determined from the complete event context.

The resulting dataset contains 10,052 video clips with an average duration of
6.1 seconds. Sampling one representative frame per clip and adding 682
event-grounded news images yields 10,734 images, as shown in Figure~\ref{fig:dataset}.

\subsection{Event-Grounded Annotation}
\label{sec:event_annotation}

Using a dedicated interface, annotators inspect each restricted visual sample
together with its source-grounded reference context: the original webpage and
headline for news/interview samples, or expanded dialogue and scene context
for television-series samples. They identify the focal person and retain
samples for which the restricted input is insufficient or potentially
misleading. Samples with low visual quality, no identifiable focal person, or
inadequate event grounding are excluded.

Each retained sample is assigned one of 11 labels: \emph{happy}, \emph{sad},
\emph{angry}, \emph{disgust}, \emph{fear}, \emph{surprise}, \emph{neutral},
\emph{pride}, \emph{embarrassed}, \emph{contempt}, or \emph{respect}. Labels
are determined from the focal person's reaction and role in the complete
event rather than facial appearance alone. The reference materials are
retained for diagnostic evaluation but are not provided in the standard EGER
setting.

\begin{wraptable}{r}{0.48\textwidth}
\centering
\caption{
Human performance with and without reference event context on 480 instances.\\
}
\label{tab:human_event_grounding}
\small
\setlength{\tabcolsep}{6pt}
\renewcommand{\arraystretch}{1.15}
\begin{tabular}{lccc}
\toprule
Evidence Condition & UAR & Acc. & WAF \\
\midrule
Visual Input
& 27.80 & 33.96 & 33.89 \\
Visual + Event Context
& \textbf{65.80} & \textbf{72.08} & \textbf{72.66} \\
\bottomrule
\end{tabular}
\end{wraptable}

\paragraph{Human validation.}
We conduct an independent study with six annotators on 480 randomly sampled
instances. For each instance, different annotators evaluate the emotion under
two evidence conditions: the restricted visual input alone, or the same input
together with its reference event context. All annotators are blinded to the
benchmark labels, and no annotator sees the same instance under both
conditions. As shown in Table~\ref{tab:human_event_grounding}, event context
raises accuracy from 33.96\% to 72.08\%, with corresponding gains in UAR and
WAF. This substantial gap supports the premise that the restricted visual
input alone does not reliably determine the emotion label.

\subsection{Visual Input Settings and Splits}
\label{sec:input_settings}
\label{sec:statistics}

We define four settings,
$\mathcal{M}=\{\text{VID-V},\text{VID-VS},\text{IMG-V},\text{IMG-VS}\}$, where \texttt{VID}/\texttt{IMG} denotes video/image and \texttt{VS} retains visible subtitles or scene text, while \texttt{V} removes detected text regions. No audio or separately supplied transcript is used. PaddleOCR~\citep{cui2025paddleocr} detects text regions and ProPainter~\citep{zhou2023propainter} inpaints the resulting masks. Each text-retained and text-removed pair shares the same underlying frames, and AffectReveal and its downstream model always receive the same version.

The benchmark contains 57\% news/interview and 43\% television-series
samples. We use an approximately $6{:}1{:}3$ train/validation/test split,
yielding 6,036/1,006/3,010 videos and 6,441/1,073/3,220 images. Representative
frames inherit their source-video assignments to prevent frame-level overlap;
the additional web images follow the same split ratio, with domain
proportions maintained across subsets.

\section{AffectReveal}
\label{sec:method}

\subsection{Overview}
\label{sec:method_overview}

\begin{figure*}[t]
    \centering
    \includegraphics[width=\textwidth]{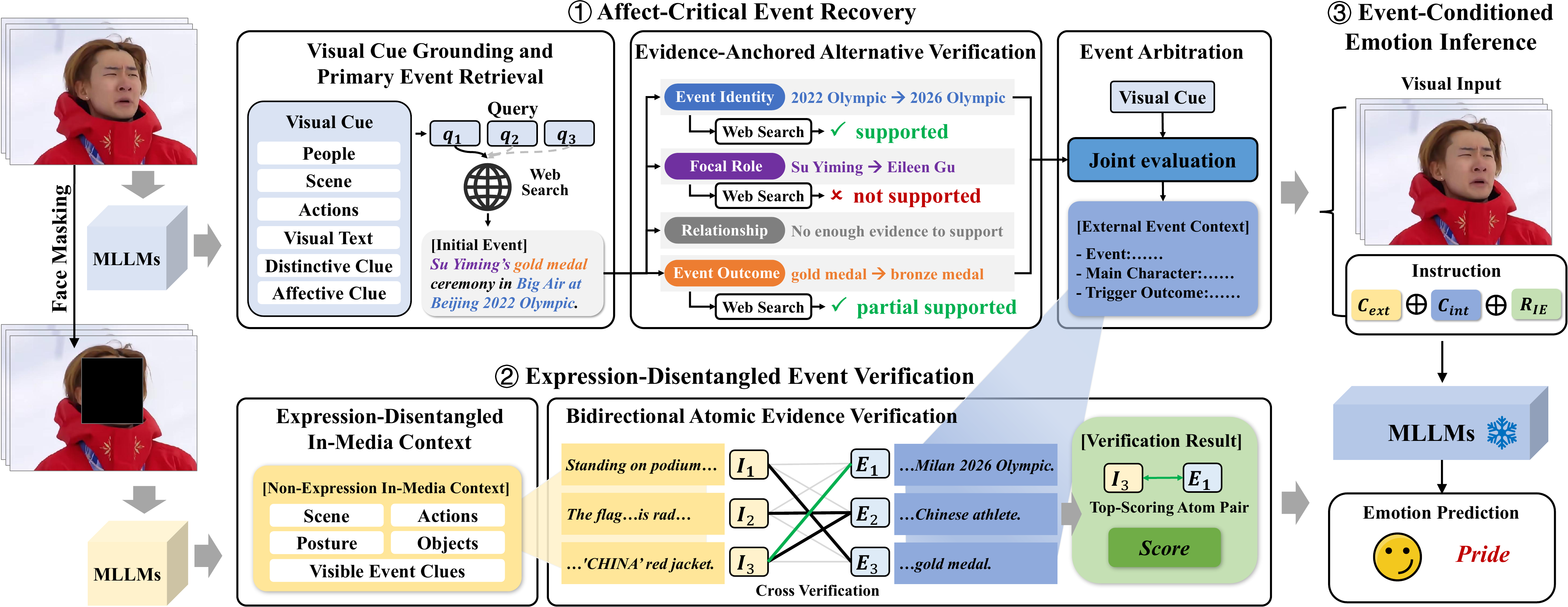}
    \vspace{-0.7cm}
    \caption{
    Overview of AffectReveal.
    The framework recovers external event context by retrieving a primary
    event and verifying evidence-anchored alternatives, then cross-checks the
    recovered event against non-facial in-media facts before emotion
    prediction.
    }
    \label{fig:overview}
\end{figure*}

EGER presents two coupled challenges.
First, limited visual evidence may support multiple event interpretations
whose differences can change the implied emotion.
We refer to this problem as \emph{affect-critical event ambiguity}.
Second, the expression that motivates event recovery may itself be
misleading; using it again to validate the recovered event creates
\emph{appearance-induced circularity}.
AffectReveal addresses these challenges through two corresponding stages,
as illustrated in Figure~\ref{fig:overview}.

First, \textbf{Affect-Critical Event Recovery} operationalizes event
ambiguity through four recurrent factors:
$
    \mathcal{D}
    =
    \{
    d_{\mathrm{id}},
    d_{\mathrm{role}},
    d_{\mathrm{rel}},
    d_{\mathrm{out}}
    \}
$,
corresponding to event identity, focal-person role, interpersonal
relationship, and event outcome.
This factorization organizes event verification rather than providing an
exhaustive definition of event context.
The stage retrieves and verifies candidate interpretations to produce an
external event context:
\begin{equation}
    C_{\mathrm{ext}}
    =
    \operatorname{Recover}(x^{m},\mathcal{K};\mathcal{D}).
\label{eq:event_recovery_overview}
\end{equation}
Second, \textbf{Expression-Disentangled Event Verification} masks facial
regions and extracts non-facial in-media context:
\begin{equation}
    C_{\mathrm{int}}
    =
    \operatorname{Extract}
    \left(
    \mathcal{M}_{\mathrm{face}}(x^{m})
    \right),
\label{eq:internal_context_overview}
\end{equation}
where $\mathcal{M}_{\mathrm{face}}$ denotes face masking.
It then evaluates the observable compatibility between the internal and
external contexts:
\begin{equation}
    R_{IE}
    =
    \operatorname{Verify}
    (C_{\mathrm{int}},C_{\mathrm{ext}}).
\label{eq:event_verification_overview}
\end{equation}
The original unmasked input, $C_{\mathrm{ext}}$, $C_{\mathrm{int}}$, and
$R_{IE}$ are jointly used for final emotion prediction.

\subsection{Affect-Critical Event Recovery}
\label{sec:event_recovery}

\paragraph{Visual Cue-Grounded Retrieval.}
To recover the external event context $C_{\mathrm{ext}}$, we first extract
structured clues from $x^{m}$, including people, scenes, actions, visible
text, distinctive objects, and observable reactions. The strongest clues are
combined into $N$ ranked event-oriented queries,
$Q=\{(q_i,p_i)\}_{i=1}^{N}$, where $p_i$ denotes query priority.
The queries and representative visual frames are used for web and image
search. Retrieved evidence is checked against the visual clues and summarized
as a primary event $e_0$, including its description, time, location,
affect-critical details, and supporting sources.

\paragraph{Alternative Verification.}
A plausible $e_0$ may still contain an incorrect event identity, focal-person
role, relationship, or outcome.
For each uncertain factor $d_i\in\mathcal{D}$, we construct an alternative
that changes only that factor:
$
    e_i^{\mathrm{alt}}
    =
    e_0[d_i\leftarrow\widetilde{d}_i].
$
The alternative $\widetilde{d}_i$ must be grounded in primary retrieval
evidence rather than freely generated.
Each alternative is independently checked through a factor-specific search
and labeled as \emph{supported}, \emph{partially supported},
\emph{not supported}, \emph{not found}, or \emph{conflicting}.
Only supported and partially supported alternatives are retained.

\paragraph{Event Arbitration.}
The primary event, retained alternatives, visual clues, and retrieved
evidence are jointly reconsidered to produce $C_{\mathrm{ext}}$.
It records the best-supported event identity, focal-person role,
relationship, outcome, and sources without directly predicting an emotion
label.

\subsection{Expression-Disentangled Event Verification}
\label{sec:cross_verification}

\paragraph{Non-Facial In-Media Context.}
To verify whether $C_{\mathrm{ext}}$ is compatible with the observed sample
without reusing the potentially misleading facial expression, we mask facial
regions frame by frame and extract $C_{\mathrm{int}}$ from the masked input.
$C_{\mathrm{int}}$ describes scenes, body actions, objects, interactions, and
visible event clues. Masking is restricted to this verification branch; the
original unmasked input remains available for final emotion prediction.

\paragraph{Bidirectional Atomic Evidence Verification.}
We decompose the internal and external contexts into atomic facts:
\begin{equation}
    \mathcal{I}=\{i_j\}_{j=1}^{n},
    \qquad
    \mathcal{E}=\{e_k\}_{k=1}^{m},
\label{eq:atomic_evidence}
\end{equation}
where each statement contains one independently assessable fact.
An evidence judge assigns every pair a support score
$M_{jk}\in\{0,0.5,1.0\}$, denoting no, partial, or direct factual support.

Rather than averaging unrelated pairs, we retain the strongest counterpart
for each fact and aggregate support in both directions:
\begin{equation}
    S_{IE}
    =
    \frac{1}{2}
    \left(
    \frac{1}{n}\sum_{j=1}^{n}\max_{k}M_{jk}
    +
    \frac{1}{m}\sum_{k=1}^{m}\max_{j}M_{jk}
    \right).
\label{eq:ie_score}
\end{equation}
The two terms measure how well the in-media facts are explained by the
recovered event and how strongly the external event is grounded in observable
evidence, respectively.

The strongest aligned pairs are retained as a textual explanation $A_{IE}$,
giving the final verification result
$R_{IE}=(S_{IE},A_{IE})$.
$R_{IE}$ measures observable compatibility rather than requiring every
external event fact to appear in the media.

\subsection{Event-Conditioned Emotion Inference}
\label{sec:event_conditioned_inference}

The external context, in-media context, and their verification relation are
serialized into a fixed-format evidence package and provided with the
original unmasked input:
\begin{equation}
    C_{\mathrm{evi}}
    =
    C_{\mathrm{ext}}
    \oplus C_{\mathrm{int}}
    \oplus R_{IE},
    \qquad
    \widehat{y}
    =
    F_{\theta}(x^{m},C_{\mathrm{evi}}),  \qquad \widehat{y}\in\mathcal{Y}
\label{eq:final_prediction}
\end{equation}
Here, $\oplus$ denotes textual serialization.
No parameter of the downstream model $F_{\theta}$ is updated.

\section{Experiments}
\label{sec:experiments}

\subsection{Experimental Setup}
\label{sec:experimental_setup}

\paragraph{Benchmark and metrics.}
We evaluate on EGER-Bench under four visual input settings: video without visible text (VID-V), video with visible text (VID-VS), image without visible text (IMG-V), and image with visible text (IMG-VS).
We report Accuracy (Acc.), Unweighted Average Recall (UAR), and Weighted Average F1-score (WAF).
Given the imbalanced emotion distribution, we use UAR as the primary metric and report Acc.\ and WAF as complementary measures.

\paragraph{Downstream emotion models.}
We evaluate AffectReveal with two general-purpose models, Qwen3.5-Omni-Plus~\citep{team2026qwen3} and Qwen2.5-Omni-7B\citep{xu2025qwen25omnitechnicalreport}, and two emotion-oriented models, Emotion-LLaMA~\citep{cheng2024emotion} and AffectGPT~\citep{lian2025affectgpt}.
External search is disabled for all downstream models.
For the locally deployable models, we additionally report variants fine-tuned on the EGER-Bench training set, denoted by FT.

\paragraph{Implementation details.}
We instantiate Affect-Critical Event Recovery with Qwen3.5-Flash~\citep{team2026qwen3}. Primary
retrieval uses one web-search call and one image-search call; alternative verification uses at most three additional
web-search calls. Facial regions are detected frame by frame using MediaPipe
Face Detection with the short-range BlazeFace model~\citep{bazarevsky2019blazeface}. In-media context
extraction, atomic decomposition, and evidence scoring are implemented with
Qwen2.5-Omni-7B. All supporting and downstream models remain frozen, and
fixed-format prompts are used. Complete prompts and inference 
configurations are provided in Appendix~\ref{app:prompts}.

\subsection{Main Results Across Models and Input Settings}
\label{sec:main_results}

We first evaluate whether AffectReveal is effective across downstream model
families and visual input settings. For each model, we compare visual-only
inference with the same model augmented by AffectReveal. For the three locally
deployable models, we further repeat this comparison after task-specific
fine-tuning on EGER-Bench, allowing us to examine whether inference-time event
evidence remains useful after parameter adaptation. As shown in Table~\ref{tab:main}, AffectReveal improves UAR, accuracy, and WAF
for every untuned downstream model under all four input settings. Averaged
across the four settings, 7.05 for Emotion-LLaMA, 5.26 for AffectGPT, and 10.53 for
Qwen2.5-Omni-7B. The improvements hold for both images and videos and remain
substantial without visible text, indicating that AffectReveal is effective
across visual formats and does not rely on textual cues being present.

\begin{table*}[t]
\centering
\caption{
Performance on EGER-Bench under four visual input settings.
FT denotes task-specific fine-tuning, with gray rows indicating fine-tuned
downstream models.
}
\label{tab:main}
\renewcommand{\arraystretch}{1.38}
\begin{adjustbox}{width=0.99\textwidth}
\begin{tabular}{l|ccc|ccc|ccc|ccc}
\toprule
\multirow{2}{*}{Method}
& \multicolumn{3}{c|}{VID-V}
& \multicolumn{3}{c|}{VID-VS}
& \multicolumn{3}{c|}{IMG-V}
& \multicolumn{3}{c}{IMG-VS} \\
& UAR & Acc. & WAF
& UAR & Acc. & WAF
& UAR & Acc. & WAF
& UAR & Acc. & WAF \\
\midrule


Emotion-LLaMA~\citep{cheng2024emotion}
& 20.42 & 20.90 & 15.69
& 21.36 & 21.99 & 16.05
& 18.87 & 19.72 & 15.36
& 18.38 & 19.63 & 15.11 \\
Emotion-LLaMA + AffectReveal
& \textbf{29.10} & \textbf{29.77} & \textbf{24.51}
& \textbf{29.13} & \textbf{29.80} & \textbf{24.38}
& \textbf{24.47} & \textbf{25.16} & \textbf{20.03}
& \textbf{24.52} & \textbf{25.25} & \textbf{19.92} \\
\rowcolor[HTML]{EFEFEF}
Emotion-LLaMA (FT)
& 26.70 & 26.61 & 19.96
& 28.38 & 28.64 & 24.88
& 23.40 & 25.12 & 18.83
& 23.48 & 25.16 & 18.93 \\
\rowcolor[HTML]{EFEFEF}
Emotion-LLaMA (FT) + AffectReveal
& \textbf{31.17} & \textbf{31.69} & \textbf{26.96}
& \textbf{32.08} & \textbf{32.69} & \textbf{28.39}
& \textbf{28.14} & \textbf{28.91} & \textbf{24.35}
& \textbf{28.73} & \textbf{29.60} & \textbf{25.41} \\
\midrule

AffectGPT~\citep{lian2025affectgpt}
& 25.63 & 25.59 & 24.44
& 25.82 & 26.08 & 24.51
& 24.60 & 24.66 & 22.76
& 24.11 & 25.06 & 23.85 \\
AffectGPT + AffectReveal
& \textbf{29.48} & \textbf{30.00} & \textbf{28.02}
& \textbf{29.93} & \textbf{30.46} & \textbf{28.51}
& \textbf{30.70} & \textbf{30.25} & \textbf{28.70}
& \textbf{31.07} & \textbf{30.87} & \textbf{29.44} \\
\rowcolor[HTML]{EFEFEF}
AffectGPT (FT)
& 25.66 & 26.06 & 24.68
& 25.42 & 26.02 & 24.28
& 25.39 & 25.28 & 23.68
& 26.02 & 25.68 & 24.53 \\
\rowcolor[HTML]{EFEFEF}
AffectGPT (FT) + AffectReveal
& \textbf{31.59} & \textbf{31.21} & \textbf{29.62}
& \textbf{31.92} & \textbf{32.05} & \textbf{30.80}
& \textbf{30.80} & \textbf{30.37} & \textbf{28.89}
& \textbf{31.27} & \textbf{31.49} & \textbf{30.32} \\
\midrule

Qwen2.5-Omni-7B~\citep{xu2025qwen25omnitechnicalreport}
& 17.11 & 17.94 & 11.13
& 21.85 & 22.52 & 17.02
& 16.66 & 18.04 & 13.77
& 18.95 & 20.37 & 16.99 \\
Qwen2.5-Omni-7B + AffectReveal
& \textbf{29.25} & \textbf{29.80} & \textbf{27.11}
& \textbf{32.83} & \textbf{33.29} & \textbf{30.61}
& \textbf{26.58} & \textbf{27.67} & \textbf{25.79}
& \textbf{28.02} & \textbf{29.13} & \textbf{27.20} \\
\rowcolor[HTML]{EFEFEF}
Qwen2.5-Omni-7B (FT)
& 25.34 & 25.45 & 21.54
& 33.19 & 32.72 & 29.50
& 22.39 & 22.48 & 19.25
& 28.88 & 28.60 & 25.90 \\
\rowcolor[HTML]{EFEFEF}
Qwen2.5-Omni-7B (FT) + AffectReveal
& \textbf{36.55} & \textbf{36.08} & \textbf{35.44}
& \textbf{36.48} & \textbf{36.78} & \textbf{33.54}
& \textbf{33.13} & \textbf{32.55} & \textbf{32.07}
& \textbf{35.94} & \textbf{35.31} & \textbf{34.86} \\
\bottomrule
\end{tabular}
\end{adjustbox}
\end{table*}

AffectReveal also remains effective after task-specific fine-tuning. Across
Emotion-LLaMA, AffectGPT, and Qwen2.5-Omni-7B, the untuned model augmented with
AffectReveal outperforms its fine-tuned visual-only counterpart in all 12
accuracy comparisons. Applying AffectReveal to the fine-tuned models yields
further gains across every model and input setting. This shows that
task-specific parameter adaptation and instance-specific event evidence are
complementary: fine-tuning improves how a model interprets the available
input, whereas AffectReveal supplies event information that the input itself
does not contain.

\subsection{Diagnostic Value of External Event Context}
\label{sec:reference_context}

To isolate the effect of missing event information, we conduct a diagnostic
experiment with Qwen3.5-Omni-Plus under VID-V. We compare visual-only
inference with two event-conditioned settings. \textbf{Reference Event
Context} provides the source-grounded event description preserved during
benchmark construction, whereas \textbf{AffectReveal} recovers event context
automatically without access to this description. Explicit emotion labels and
direct emotion-descriptive statements are removed from the reference context.
Because it may still omit relevant role, relationship, or outcome details, it
is treated as a diagnostic reference rather than a perfect oracle.

As shown in Table~\ref{tab:compact_analyses}(a), reference event context
raises UAR from 24.15\% to 39.36\%, confirming that missing event information
constitutes a substantial bottleneck. AffectReveal reaches 34.76\% UAR using
automatically recovered evidence, closing 69.8\% of the gap between the
visual-only and reference-context conditions.
Similar trends are observed for UAR and WAF. These results show that
AffectReveal recovers a substantial portion of the task-relevant event
information without privileged access to the benchmark reference context.

\begin{table*}[t]
\centering
\caption{
Diagnostic and ablation results under VID-V.
(a) Event-context diagnostic using Qwen3.5-Omni-Plus;
(b) cumulative component analysis and
(c) verification ablations using Qwen2.5-Omni-7B.
UAR is the primary metric.
}
\label{tab:compact_analyses}

\fontsize{7.5pt}{8.5pt}\selectfont
\setlength{\tabcolsep}{3pt}
\renewcommand{\arraystretch}{1.22}

\begin{minipage}[t]{0.33\textwidth}
\vspace{0pt}
\centering
\textbf{(a) Event Context}\\[2pt]
\begin{tabularx}{\linewidth}{@{}Xcc@{}}
\toprule
Evidence Setting & UAR & WAF \\
\midrule
Visual Input Only
& 24.15 & 20.35 \\
Visual + Reference Event
& \textbf{39.36} & \textbf{39.02} \\
Visual + AffectReveal
& 34.76 & 33.32 \\
\bottomrule
\end{tabularx}
\end{minipage}
\hfill
\begin{minipage}[t]{0.36\textwidth}
\vspace{0pt}
\centering
\textbf{(b) Evidence Components}\\[2pt]
\begin{tabularx}{\linewidth}{@{}Xcc@{}}
\toprule
Evidence Setting & UAR & WAF \\
\midrule
Visual Input Only
& 17.11 & 11.13 \\
+ External Event Context $C_{\mathrm{ext}}$
& 27.84 & 25.73 \\
+ In-Media Context $C_{\mathrm{int}}$
& 27.75 & 25.56 \\
+ AffectReveal $R_{\mathrm{IE}}$
& \textbf{29.25} & \textbf{27.11} \\
\bottomrule
\end{tabularx}
\end{minipage}
\hfill
\begin{minipage}[t]{0.28\textwidth}
\vspace{0pt}
\centering
\textbf{(c) Verification Designs}\\[2pt]
\begin{tabularx}{\linewidth}{@{}Xcc@{}}
\toprule
Variant & UAR & WAF \\
\midrule
w/o Face Masking
& 28.89 & 24.37 \\
w/o Atomic Support
& 28.95 & 26.74 \\
Full AffectReveal
& \textbf{29.25} & \textbf{27.11} \\
\bottomrule
\end{tabularx}
\end{minipage}
\end{table*}

\subsection{Does Affect-Critical Verification Improve Event Recovery?}
\label{sec:retrieval_analysis}

Retrieving a relevant event does not necessarily recover the details needed
for emotion interpretation. We therefore compare three strategies:
\textbf{Primary Retrieval}, which uses only the initial search;
\textbf{Generic Multi-Search}, which performs unconstrained query refinement;
and \textbf{Affect-Critical Event Recovery}, which verifies evidence-grounded
alternatives that differ in one affect-critical factor. Generic Multi-Search
and our method use the same maximum retrieval budget, isolating the effect of
how additional searches are organized.

We evaluate the recovered context against the source-grounded reference event
at both event and factor levels. A recovery is labeled \emph{Match} when the
central event and its affect-critical details agree with the reference,
\emph{Partial} when the central event is recovered but some details remain
incomplete or unresolved, and \emph{Mismatch} when the event or an
affect-determining detail conflicts with the reference. We additionally
measure correctness for event identity, focal-person role, interpersonal
relationship, and event outcome. The complete evaluation protocol and its
agreement with human judgments are provided in
Appendix~\ref{app:event_evaluation}.

As shown in Table~\ref{tab:event_recovery}, Primary Retrieval produces a
fully or partially aligned event in 34.80\% of the samples. Generic
Multi-Search increases this rate to 38.88\%, showing that additional retrieval
itself is beneficial. Under the same retrieval budget, Affect-Critical Event
Recovery further raises it to 42.86\% and reduces the mismatch rate to
57.14\%. The largest factor-level gain over Generic Multi-Search occurs for
interpersonal relationship, increasing from 25.53\% to 31.06\%, with
additional gains for event identity, focal-person role, and event outcome.
These results show that organizing retrieval around affect-critical
alternatives recovers emotion-relevant event details more reliably than
unconstrained query refinement.

\begin{table*}[t]
\centering
\caption{
Event recovery quality under VID-V. Generic Multi-Search and Affect-Critical
Event Recovery use the same maximum retrieval budget.
}
\label{tab:event_recovery}
\fontsize{8pt}{9pt}\selectfont
\setlength{\tabcolsep}{2pt}
\renewcommand{\arraystretch}{1.42}
\begin{tabular}{l|cccc|ccc}
\toprule
\multirow{2}{*}{Method}
& \multicolumn{4}{c|}{Affect-Critical Factor Correctness (\%) $\uparrow$}
& \multicolumn{3}{c}{Overall Event Recovery (\%)} \\
& Event Identity
& Focal Role
& Relationship
& Event Outcome
& Match $\uparrow$
& Partial $\uparrow$
& Mismatch $\downarrow$ \\
\midrule
Primary Retrieval
& 16.35 & 36.67 & 23.33 & 12.25
& 13.25 & 21.55 & 65.20 \\
Generic Multi-Search
& 15.44 & 38.90 & 25.53 & 14.32
& 12.96 & 25.92 & 61.12 \\
\textbf{Affect-Critical Event Recovery}
& \textbf{16.52} & \textbf{40.64} & \textbf{31.06} & \textbf{15.03}
& \textbf{13.95} & \textbf{28.91} & \textbf{57.14} \\
\bottomrule
\end{tabular}
\end{table*}

\begin{figure*}[t]
    \centering
    \includegraphics[width=\textwidth]{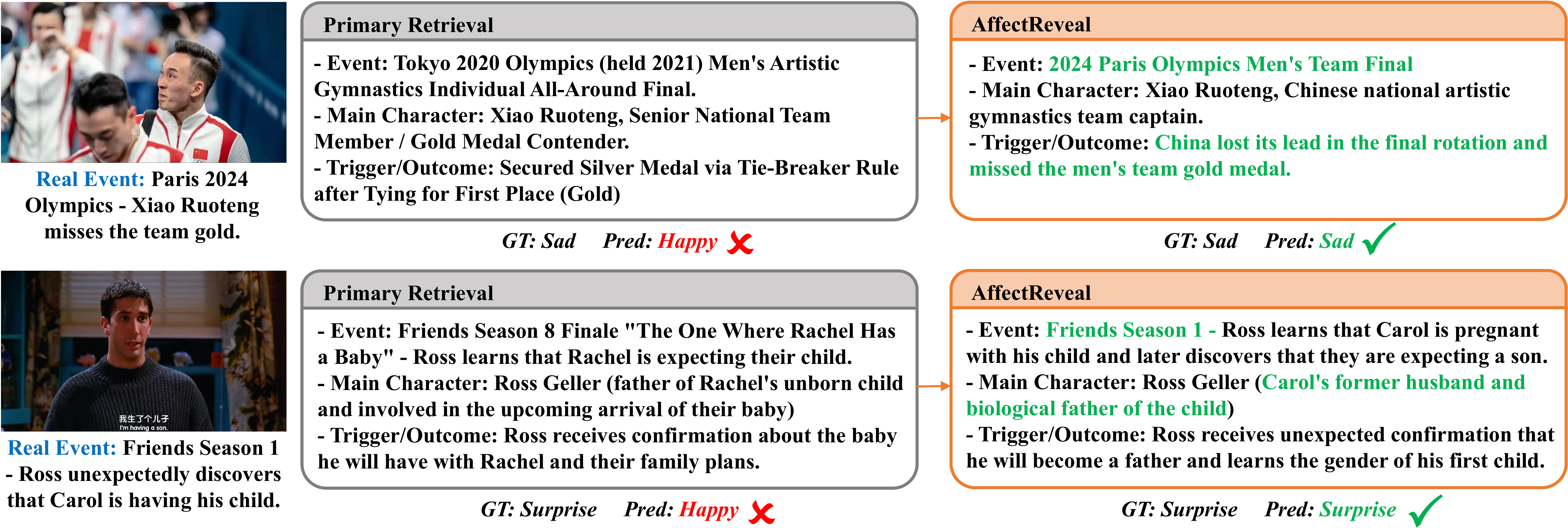}
    \vspace{-0.6cm}
    \caption{
    Qualitative comparisons between Primary Retrieval and AffectReveal.
    Although Primary Retrieval identifies the correct focal person, it
    retrieves a topically related but affectively incorrect event.
    AffectReveal corrects the event identity and outcome in the Olympic
    example, and the event identity and interpersonal relationship in the
    television-series example, leading to reference-aligned emotion
    predictions. Green text highlights the corrected event details.
    }
    \label{fig:visualization}
\end{figure*}

\subsection{How Does Event Verification Contribute to Emotion Recognition?}
\label{sec:verification_analysis}

The recovered external context may be incomplete or partially mismatched with
the observed sample. We therefore examine whether in-media facts should be
treated simply as additional context or used to assess the compatibility of
the recovered event. We first incrementally add the external event context
$C_{\mathrm{ext}}$, the expression-disentangled in-media context
$C_{\mathrm{int}}$, and their explicit support relation $R_{IE}$.
As shown in Table~\ref{tab:compact_analyses}(b), external event context
provides the dominant gain, raising UAR from 17.11\% to 27.84\%. Directly
adding in-media context provides no further improvement, whereas explicitly
modeling its support relation with the external event raises UAR to 29.25\%
and WAF to 27.11\%. This indicates that
in-media facts are more useful for assessing the recovered event than as
another unconstrained textual description.

We next ablate the two mechanisms used to construct this support relation. The
\emph{w/o face masking} variant extracts $C_{\mathrm{int}}$ from the original
input, allowing the facial expression being interpreted to also participate
in event verification. The \emph{w/o atomic support} variant replaces
fact-level decomposition and alignment with a holistic comparison of the two
contexts.
As shown in Table~\ref{tab:compact_analyses}(c), removing face masking
reduces WAF by 2.74 points, supporting the concern that a potentially
misleading expression can introduce circular evidence during verification.
Replacing atomic support with holistic comparison causes smaller but
consistent decreases across all metrics. Together, these results show
that event verification is most effective when it uses expression-disentangled
in-media facts and compares them with external evidence at the level of
individual factual statements.

\subsection{Qualitative Analysis}
\label{sec:qualitative}

Figure~\ref{fig:visualization} illustrates that recognizing the focal person
or retrieving a related event is insufficient for EGER. In the Olympic
example, Primary Retrieval identifies Xiao Ruoteng but associates him with
the Tokyo individual all-around final and a silver-medal outcome, leading to
a prediction of happiness. AffectReveal instead recovers the Paris team final
and the missed opportunity for gold, supporting sadness. In the television
example, Primary Retrieval identifies Ross but confuses Carol's pregnancy
with a later storyline involving Rachel. Correcting the event identity and
interpersonal relationship changes the prediction from happiness to
surprise. These cases demonstrate that AffectReveal improves emotion
recognition by resolving affect-critical event details rather than merely
retrieving information about the visible person.

\section{Conclusion}

We introduced \textbf{Event-Grounded Emotion Recognition} and
\textbf{EGER-Bench} for recognizing visual reactions whose affect-determining
events are not fully observable. We also proposed \textbf{AffectReveal},
which verifies factorized event alternatives and cross-checks recovered
context against expression-disentangled visual evidence. Across multiple
models and visual settings, AffectReveal consistently improves recognition
and complements task-specific fine-tuning without updating downstream
parameters. These results highlight the importance of acquiring and verifying
the events behind visually ambiguous reactions. Future work will extend EGER
to dynamic and interactive settings where systems actively seek evidence and
revise emotion interpretations as events unfold.

\section*{AI use statement}

Generative AI tools were used during manuscript preparation to assist with language polishing, text compression, and \LaTeX{} formatting.
The authors reviewed and revised all AI-assisted content and independently verified the technical claims, citations, experimental results, and numerical analyses reported in the paper.
Generative AI was not used as an authoritative source for scientific claims or references, nor was it used to fabricate experimental data or results.
The authors take full responsibility for the final content of the submission.

\section*{Ethics statement}

Emotion recognition from human behavior raises ethical concerns related to privacy, subjective interpretation, cultural variation, and potential misuse in surveillance or high-stakes decision-making.
EGER-Bench is intended solely for research on event-grounded emotion understanding and should not be interpreted as providing objective psychological assessments of individuals.
EGER-Bench is constructed from news/interview media and television-series content with source provenance retained for event grounding.
For any release, we will respect applicable source licenses and copyright restrictions and provide documentation describing the intended research use and limitations of the benchmark.
We do not advocate the use of AffectReveal or EGER-Bench for surveillance, clinical assessment, employment screening, law enforcement, or other high-stakes decisions about individuals.

\section*{Reproducibility statement}

To facilitate reproducibility, Sections~3--5 describe the dataset construction, AffectReveal framework, and experimental settings. The appendix provides complete prompts, evaluation protocols, implementation details, and inference configurations used in our experiments.

\bibliography{ref}
\bibliographystyle{iclr2027_conference}

\newpage

\appendix

\section*{Appendix Contents}

The appendix is structured as follows:

\begin{itemize}[leftmargin=1.5em]
    \item Appendix~\ref{app:qwen35} reports additional EGER-Bench results with Qwen3.5-Omni-Plus, an API-accessed closed-source downstream model.

    \item Appendix~\ref{app:mer2023} presents a vision-only transfer evaluation on MER2023, examining whether recovered event evidence can also benefit conventional visual emotion recognition.

    \item Appendix~\ref{app:event_evaluation} validates the automatic event recovery evaluation by comparing Qwen-based assessments with human annotations.

    \item Appendix~\ref{app:source_analysis} analyzes AffectReveal's performance across News/Interview and TV Series source domains.

    \item Appendix~\ref{app:prompts} provides the prompt templates and inference configuration used by AffectReveal's core components.
\end{itemize}

\section{Additional Results with Qwen3.5-Omni-Plus}
\label{app:qwen35}

In addition to the locally deployable downstream models reported in the main comparison, we evaluate AffectReveal with Qwen3.5-Omni-Plus, an API-accessed closed-source vision-capable MLLM. Since its model parameters are not available, we evaluate the untuned model only and do not include a task-specific fine-tuning variant. External search is disabled for the downstream model itself, following the same protocol as the main experiments; external evidence is introduced only through AffectReveal.

\begin{table*}[h]
\centering
\caption{
Performance of Qwen3.5-Omni-Plus on EGER-Bench under four visual input settings.
}
\label{tab:qwen35_appendix}
\renewcommand{\arraystretch}{1.38}
\begin{adjustbox}{width=0.99\textwidth}
\begin{tabular}{l|ccc|ccc|ccc|ccc}
\toprule
\multirow{2}{*}{Method}
& \multicolumn{3}{c|}{VID-V}
& \multicolumn{3}{c|}{VID-VS}
& \multicolumn{3}{c|}{IMG-V}
& \multicolumn{3}{c}{IMG-VS} \\
& UAR & Acc. & WAF
& UAR & Acc. & WAF
& UAR & Acc. & WAF
& UAR & Acc. & WAF \\
\midrule

Qwen3.5-Omni-Plus~\citep{team2026qwen3}
& 24.15 & 25.00 & 20.35
& 34.91 & 35.56 & 33.24
& 20.35 & 22.37 & 17.71
& 27.08 & 28.98 & 26.01 \\

Qwen3.5-Omni-Plus + AffectReveal
& \textbf{34.76} & \textbf{35.41} & \textbf{33.32}
& \textbf{39.50} & \textbf{40.11} & \textbf{38.30}
& \textbf{29.16} & \textbf{30.84} & \textbf{28.77}
& \textbf{31.70} & \textbf{33.25} & \textbf{31.45} \\

\bottomrule
\end{tabular}
\end{adjustbox}
\end{table*}

As shown in Table~\ref{tab:qwen35_appendix}, AffectReveal consistently improves Qwen3.5-Omni-Plus across all four visual input settings and all three evaluation metrics. The average UAR improvement across the four settings is 7.16 points. The largest UAR gain occurs under VID-V, increasing from 24.15 to 34.76, while IMG-V improves from 20.35 to 29.16. Improvements are also retained when visible text is available, with UAR increasing from 34.91 to 39.50 under VID-VS and from 27.08 to 31.70 under IMG-VS.

These results complement the locally deployable models reported in the main experiments and show that the benefit of AffectReveal is not restricted to downstream models whose parameters or training procedures are accessible.

\section{Vision-Only Transfer to MER2023}
\label{app:mer2023}
As a secondary transfer analysis, we evaluate the same inference-time framework on the vision-only setting of MER2023.  Table~\ref{tab:mer2023} reports the corresponding results. This experiment is not used to establish the EGER task because MER2023 does not explicitly require external event context; it instead tests whether recovered event evidence can help conventional visual emotion recognition when the observed media is limited.

\begin{table}[h]
\centering
\caption{Vision-only transfer results on MER2023.}
\label{tab:mer2023}
\fontsize{8pt}{9pt}\selectfont
\setlength{\tabcolsep}{7pt}
\renewcommand{\arraystretch}{1.35}
\begin{tabular}{lccc}
\toprule
Method & Acc. & UAR & WAF \\
\midrule
Qwen3.5-Omni-Plus & 78.63 & \textbf{69.66} & 79.20 \\
Qwen3.5-Omni-Plus+AffectReveal & \textbf{81.51} & 69.60 & \textbf{81.61} \\
\midrule
Emotion-LLaMA & 42.81 & 32.71 & 36.60 \\
Emotion-LLaMA+AffectReveal & \textbf{50.96} & \textbf{38.44} & \textbf{48.18} \\
\midrule
AffectGPT & 48.44 & 20.57 & 49.47 \\
AffectGPT+AffectReveal & \textbf{53.84} & \textbf{23.69} & \textbf{56.73} \\
\midrule
Qwen2.5-Omni-7B & 36.69 & 37.64 & 36.86 \\
Qwen2.5-Omni-7B+AffectReveal & \textbf{41.25} & \textbf{45.54} & \textbf{44.24} \\
\bottomrule
\end{tabular}
\end{table}

\section{Validation of Automatic Event Recovery Evaluation}
\label{app:event_evaluation}
\begin{figure*}[t]
    \centering
    \includegraphics[width=0.4\textwidth]{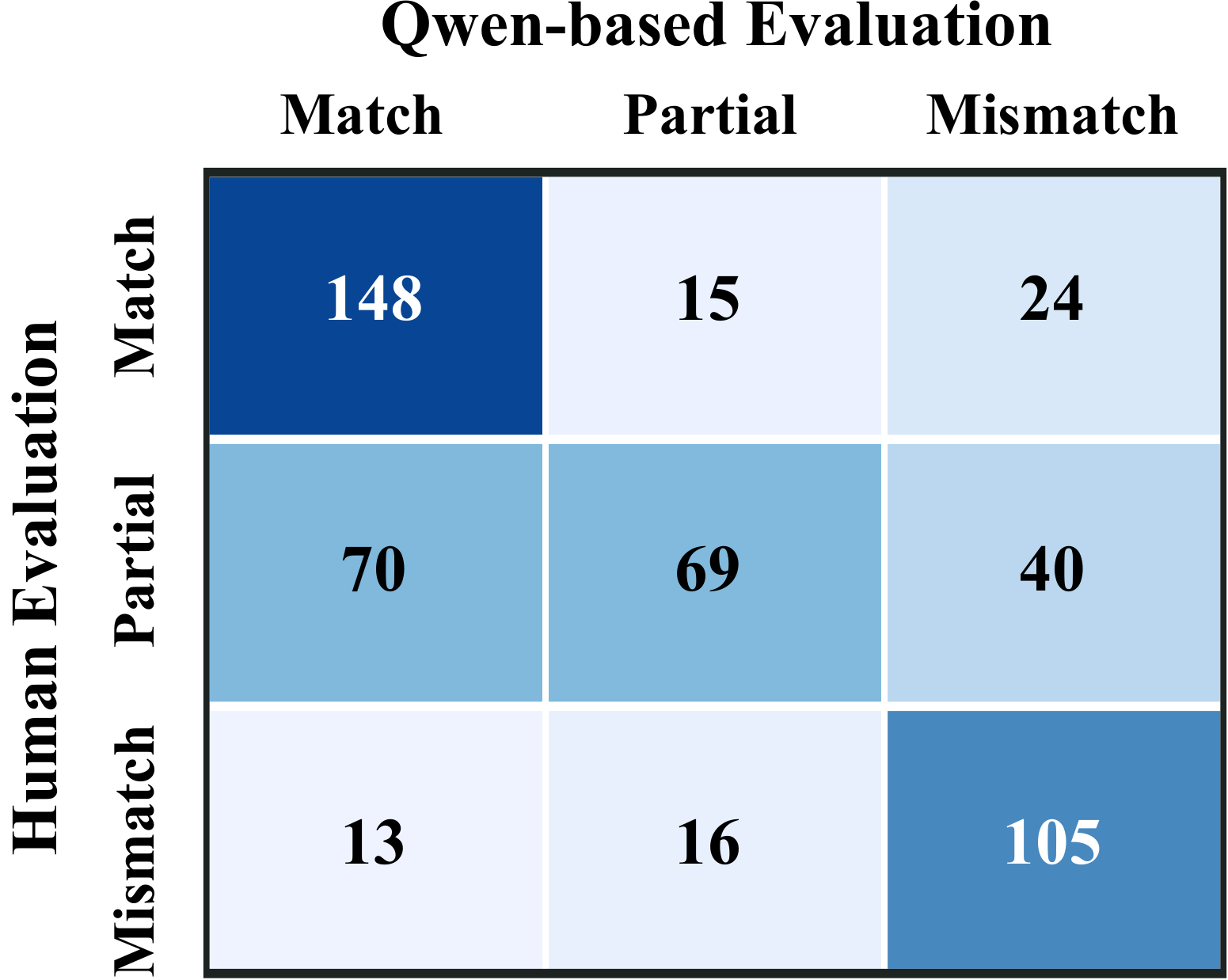}
    \caption{Comparison between Qwen-based and human assessments of event recovery quality. The confusion matrix is computed on 500 randomly sampled VID-VS test samples using three recovery levels: Match, Partial, and Mismatch.}
    \label{fig:human_event}
\end{figure*}

To analyze the correspondence between automatic and human assessment of event recovery quality, we compare Qwen-based judgments with human annotations on a randomly sampled subset. Specifically, we randomly select 500 test samples under the VID-VS setting and ask six annotators to categorize each recovered event context into three levels: Match, Partial Match, and Mismatch. Figure~\ref{fig:human_event} presents the confusion matrix between Qwen-based and human assessments. The two evaluations agree at 64.4\%, with most samples concentrated along the diagonal. The remaining differences mainly occur between neighboring categories, especially for partially recovered events, reflecting the ambiguity in assessing event-level consistency.

\section{Performance Across Source Domains}
\label{app:source_analysis}

EGER-Bench contains two source domains with different forms of event context.
News and interview samples are typically associated with publicly documented events, whereas television-series samples depend more heavily on narrative context, character roles, and interpersonal relationships.
We therefore examine whether AffectReveal improves upon initial event retrieval in both domains.

We partition the VID-V test set into News/Interview and TV Series subsets and evaluate Qwen2.5-Omni-7B separately on each.
\textbf{Primary Retrieval} conditions the downstream model on the event interpretation obtained from the initial search, whereas \textbf{AffectReveal} applies the complete event recovery and cross-verification pipeline.

\begin{table*}[h]
\centering
\caption{
Emotion recognition across source domains under VID-V using Qwen2.5-Omni-7B. News/Interview and TV Series account for 57\% and 43\% of the evaluation samples, respectively.
}
\label{tab:source_breakdown}
\fontsize{8pt}{9pt}\selectfont
\setlength{\tabcolsep}{7pt}
\renewcommand{\arraystretch}{1.42}
\begin{tabular}{l|ccc|ccc|ccc}
\toprule
\multirow{2}{*}{Method} & \multicolumn{3}{c|}{New/Interview (57\%)} & \multicolumn{3}{c|}{TV Series (43\%)} & \multicolumn{3}{c}{All} \\
 & UAR & Acc. & WAF & UAR & Acc. & WAF & UAR & Acc. & WAF \\
 \midrule
Primary Retrieval & 28.87 & 32.40 & 27.23 & 23.64 & 23.92 & 23.67 & 28.09 & 28.74 & 25.09 \\
\textbf{AffectReveal} & \textbf{29.45} & \textbf{32.69} & \textbf{27.99} & \textbf{25.24} & \textbf{26.00} & \textbf{27.45} & \textbf{29.25} & \textbf{29.80} & \textbf{27.11} \\ \bottomrule
\end{tabular}
\end{table*}

As shown in Table~\ref{tab:source_breakdown}, AffectReveal improves all three metrics in both source domains.
On News/Interview, it improves accuracy, UAR, and WAF over Primary Retrieval by 0.29, 0.58, and 0.76 points, respectively.
The gains are larger on TV Series, reaching 2.08 points in accuracy, 1.60 in UAR, and 3.78 in WAF.
Across all samples, AffectReveal improves accuracy from 28.74\% to 29.80\%, UAR from 28.09\% to 29.25\%, and WAF from 25.09\% to 27.11\%.

These results show that AffectReveal provides consistent benefits across both publicly documented and narrative-driven events.
Its larger improvement on television-series samples is consistent with the motivation of resolving affect-critical ambiguity, as these samples more often require information about narrative outcomes, character roles, and interpersonal relationships beyond the focal visual input.

\section{Prompt Templates and Inference Configuration}
\label{app:prompts}

This section lists the fixed prompts used by the core components of AffectReveal together with the corresponding inference settings. Following the presentation style of agentic systems, we represent sample-dependent inputs with brace-enclosed placeholders, while keeping the fixed instructions unchanged from the implementation.

\paragraph{Inference Configuration.}
For Qwen3.5-Flash, thinking is disabled for the initial visual analysis within Visual Cue-Grounded Retrieval, enabled for the tool-using primary retrieval and Alternative Verification, and disabled for Event Arbitration and the final serialization of the recovered external event context $C_{\mathrm{ext}}$. Tool selection is automatic during the retrieval calls.
For Qwen2.5-Omni-7B, which is used for Non-Facial In-Media Context and Bidirectional Atomic Evidence Verification, we use greedy decoding with \texttt{do\_sample=False} and \texttt{max\_new\_tokens=512}. Video input for Non-Facial In-Media Context is sampled at 1 FPS with audio disabled.

\subsection{Affect-Critical Event Recovery}
\label{app:prompts_recovery}

\paragraph{Visual Cue-Grounded Retrieval.}
The following is the complete video prompt used to extract retrieval-oriented visual cues and search queries in the reported visual-only setting.

\begin{lstlisting}[style=affectprompt]
VISUAL-ONLY MODE: Use visual information only. The supplied video has no audio track. Keep all audio fields empty and never use audio clues.

Analyze the video across multiple moments for an external-information retrieval agent that will later assist an emotion classification model.

Do not identify a final emotion label. Separate directly observed facts from hypotheses. Extract only clues useful for identifying the person, event, place, relationship, or outcome.

Return exactly one JSON object with this schema:
{
  "observed": {
    "people": ["brief visible description"],
    "scene": "brief scene description",
    "actions": ["important temporal action or interaction"],
    "visible_text": ["OCR/logo/name/number if readable"],
    "distinctive_clues": ["clothing, object, venue, organization, broadcast or event clue"],
    "affective_cues": ["visible tears, smile, restrained posture, cheering, embrace, etc."],
    "uncertainties": ["important unclear visual point"]
  },
  "queries": [
    {
      "query": "concise web-search query",
      "purpose": "identify_event|identify_person|verify_fact",
      "basis": "objective|hypothesis"
    }
  ],
  "visual_search_recommended": true
}

Rules:
1. Integrate the whole video rather than describing every frame.
2. Produce 2 queries by default and no more than 3.
3. Query 1 should be the most reliable objective query.
4. Always set visual_search_recommended=true.
5. Do not output a final emotion label, Markdown, or text outside JSON.
\end{lstlisting}

The structured visual analysis is then used for primary event retrieval and evidence-grounded alternative construction.

\begin{lstlisting}[style=affectprompt]
You are the primary retrieval module for a hidden-emotion understanding agent.
Your first task is to identify the most likely real-world event. Your second task is NOT to invent counterfactual stories, but to harvest a small set of evidence-anchored alternative event interpretations from the material you actually retrieved.

Stage-1 visual analysis:
{stage1_result_json}

Tool policy:
1. Perform ONE focused web_search using the strongest objective combination of
   visual, OCR, entity, number, place, and event clues.
2. Representative visual input is attached. Perform exactly ONE image_search for same/near-duplicate matching.
3. Use web_extractor only for up to 3 promising pages
   whose body is needed to verify identity, role, event outcome, relationship, or
   a concrete event title/description.
4. Prefer reputable/directly matching sources. Preserve contradictions.
5. Do not search generic emotion knowledge. Search the concrete event background.

STEP A — Build a PRIMARY EVENT ANCHOR.
Extract the stable event core from retrieved evidence: named people/teams/organization, place, date/time, competition/program/incident, score/result, quoted phrase, article headline, or exact/near-duplicate image match. Separate verified and uncertain slots.

STEP B — Harvest EVIDENCE-ANCHORED COUNTERFACTUAL CANDIDATES.
Generate 0 to 3 alternatives ONLY if the retrieved material itself supplies a concrete reason for that alternative. A valid candidate must:
- share the primary event core or be a directly confusable nearby real event;
- change exactly ONE uncertainty slot: subject_role, event_outcome, relationship, or event_identity;
- cite at least 1 concrete evidence_anchor_facts from the retrieved search result/snippet/extracted page/image match;
- remain compatible with Stage-1 directly observed visual facts;
- be specific enough to verify with a focused web query.

IMPORTANT:
- Do NOT create an alternative merely because it is semantically possible.
- Do NOT use a generic opposite such as win->lose, parent->coach, award->elimination unless the retrieved material mentions or strongly implies that concrete alternative.
- Do NOT force a fixed number of candidates. Returning [] is correct when retrieval exposes no grounded alternative.
- Prefer a candidate mentioned by a second retrieved article/title/snippet over a candidate invented from world knowledge.

Return exactly one JSON object:
{
  "primary_query_used": "main text query used for web_search",
  "best_event": "current best identified event, or not reliably identified",
  "best_event_confidence": 0.0,
  "image_match_strength": "exact|near_duplicate|similar|none|not_used",
  "event_anchor": {
    "entities": ["person/team/organization explicitly supported"],
    "event_name": "event/program/match/incident name or uncertain",
    "time_place": "date/place if supported",
    "stable_facts": ["facts shared by plausible interpretations"],
    "uncertain_slots": ["subject_role|event_outcome|relationship|event_identity"]
  },
  "subject_identity_role": "current best role interpretation",
  "trigger_outcome": "current best result/trigger",
  "relationship_social_context": "current best relationship/social context",
  "verified_facts": ["facts directly supported by retrieved sources"],
  "candidate_facts": ["plausible but incompletely verified claims"],
  "contradictions": ["important mismatch or uncertainty"],
  "counterfactual_candidates": [
    {
      "id": "CF1",
      "dimension": "subject_role|event_outcome|relationship|event_identity",
      "priority": 1,
      "primary_event": "primary event interpretation",
      "alternative_event": "one concrete competing event interpretation",
      "shared_anchor_facts": ["facts preserved from the primary event"],
      "evidence_anchor_facts": ["retrieved fact that motivated this alternative"],
      "evidence_grounding": "strong|medium|weak|none",
      "changed_slot_primary": "value under the primary interpretation",
      "changed_slot_alternative": "value under the alternative interpretation",
      "why_plausible": "why this is a real competing interpretation under the retrieved evidence",
      "why_emotion_critical": "why this one changed slot matters for later affect interpretation",
      "expected_discriminator": "specific fact that can distinguish the two",
      "verification_query": "focused open-web query for this concrete alternative"
    }
  ]
}

Return no more than 3 candidates. Never pad the list.
Do not output a final emotion label. Do not output Markdown or text outside JSON.
\end{lstlisting}

\paragraph{Alternative Verification.}
Each evidence-grounded alternative is independently checked with the following prompt.

\begin{lstlisting}[style=affectprompt]
You are an independent verification branch for one evidence-anchored competing event interpretation. The candidate already came from retrieved evidence; your job is to verify it, not to elaborate it into a new story.

Stage-1 multimodal analysis:
{stage1_result_json}

Primary retrieval result:
{primary_result_json}

Candidate to verify:
{candidate_json}

Perform exactly ONE focused web_search using plan.verification_query or a stricter equivalent. Do not repeat the broad primary query. image_search is not available.
Use web_extractor only for up to 2 pages when body text is needed.

Verification policy:
1. Stage-1 directly observed visual facts and plan.shared_anchor_facts are invariants. The candidate may change only plan.dimension.
2. First verify that plan.evidence_anchor_facts were not misread or taken out of context. A candidate collapses if its motivating evidence is itself incorrect.
3. Prefer facts with high specificity: exact name+role, score/result, relation, quoted statement, date/place, event title, official roster/report, or a news description tied to the same event.
4. Generic overlap (crying, hugging, interview, crowd, celebration) is weak evidence.
5. Explicitly assess media_consistency and evidence_specificity.
6. Do not invent a third interpretation. Only compare the supplied alternative with the primary interpretation.

Return exactly one JSON object:
{
  "counterfactual_id": "{candidate_id}",
  "dimension": "{candidate_dimension}",
  "counterfactual_query_used": "actual query",
  "primary_event": "primary interpretation",
  "alternative_event": "candidate interpretation",
  "counterfactual_status": "supported|partially_supported|not_supported|not_found|conflicting",
  "media_consistency": "high|medium|low|none",
  "evidence_specificity": "high|medium|low|none",
  "supporting_facts": ["specific facts supporting the candidate"],
  "refuting_facts": ["specific facts refuting it or supporting primary"],
  "discriminative_facts": ["facts that distinguish the two"],
  "effect_on_primary": "strengthens_primary|weakens_primary|ambiguous|unrelated",
  "confidence": 0.0,
  "summary": "concise evidence-based comparison"
}

Do not output a final emotion label. Do not output Markdown or text outside JSON.
\end{lstlisting}

\paragraph{Event Arbitration.}
Reconsider the primary interpretation and retained alternatives together without further browsing.

\begin{lstlisting}[style=affectprompt]
Arbitrate ONE most likely event interpretation from a primary event and independently verified evidence-anchored alternatives. Do not browse and do not add facts.

Stage-1 multimodal analysis:
{stage1_result_json}
Primary event:
{primary_result_json}
Retained competing candidates:
{retained_candidates_json}
All branch results (including rejected ones for audit):
{counterfactual_results_json}
Tool counts:
{tool_counts_json}
Source records:
{source_records_json}

Arbitration policy:
1. Do not use majority voting.
2. Prefer the interpretation that jointly maximizes: media consistency, source specificity, direct event/role/outcome evidence, and independent support.
3. A candidate that needs changing more than its declared dimension is invalid.
4. Evidence compatible with both interpretations is non-discriminative and should not decide the winner.
5. If no interpretation is sufficiently discriminated, return uncertain/conflicting instead of forcing a winner.
6. Keep the final output focused on objective event, role, relationship, and outcome; do not assign an emotion label.

Return exactly one JSON object:
{
  "status": "confirmed|probable|uncertain|conflicting|not_found",
  "confidence": 0.0,
  "source_agreement": "high|medium|low|none",
  "image_match_strength": "exact|near_duplicate|similar|none|not_used",
  "causal_discrimination": "high|medium|low|none",
  "best_event": "single selected event or not reliably identified",
  "verified_facts": ["facts retained after comparison"],
  "main_character": "focal person's identity/role if supported",
  "event_trigger_outcome": "result/trigger if supported",
  "relationship_context": "relationship if supported",
  "counterfactual_test": "brief account of why alternatives were retained or rejected",
  "conflicts": ["remaining uncertainty"]
}
Do not output Markdown or text outside JSON.
\end{lstlisting}

The arbitration output is serialized into the compact external event context used by the subsequent verification stage.

\begin{lstlisting}[style=affectprompt]
Convert the materials below into a minimal JSON object for an emotion-understanding model.
Do not browse and do not add facts absent from the materials.

Current counterfactual output mode: arbitrate

Stage-1 media analysis:
{stage1_result_json}

Retrieval evidence note:
{retrieval_note}

Actual tool counts:
{tool_counts_json}

Actual source records:
{source_records_json}

ARBITRATION MODE:
- Use only the single BEST_MATCH selected by the arbitration note.
- Prefer a compact structure inside the context string: "Event: ... Main Character: ..."
- Include objective role/relationship and trigger/outcome; avoid unnecessary affective interpretation and never assign the final emotion label.

Return exactly this JSON schema:
{
  "status": "confirmed|probable|uncertain|conflicting|not_found",
  "confidence": 0.0,
  "source_agreement": "high|medium|low|none",
  "image_match_strength": "exact|near_duplicate|similar|none|not_used",
  "context": "one compact English string"
}

Context-writing rules:
1. context must never be empty, even when retrieval fails.
2. Write 45-110 English words when enough material exists.
3. Prioritize concrete event identity, main-character role, relationship, and outcome.
4. Never introduce an event/candidate absent from the retrieval note.
5. Do not mention URLs, tools, numerical scores, retrieval rounds, or source counts.
6. If identity/event is uncertain, say so explicitly rather than inventing a story.
7. Do not assign one final emotion-class label.
8. In multi_event mode, ambiguity is intentional: reliability refers to the candidate set, not to uniqueness of the true event.

Do not output Markdown or text outside JSON.
\end{lstlisting}

\subsection{Expression-Disentangled Event Verification}
\label{app:prompts_verification}

All Qwen2.5-Omni-7B calls in this stage use the following system message.

\begin{lstlisting}[style=affectprompt]
You are a multimodal context analysis assistant.
\end{lstlisting}

\paragraph{Non-Facial In-Media Context.}
After facial regions are masked, the following video prompt extracts only non-facial visual information.

\begin{lstlisting}[style=affectprompt]
The input video has masked human faces.

Analyze ONLY non-facial visual information.

Describe:
- human actions
- body posture
- objects
- environment
- human interactions
- visible event clues

Do NOT infer emotion.
Do NOT use facial expressions.
Do NOT use audio.
Do NOT use subtitles.

Return JSON:

{
 "internal_context":"..."
}
\end{lstlisting}

\paragraph{Bidirectional Atomic Evidence Verification.}
The internal and external contexts are first decomposed into atomic factual evidence.

\begin{lstlisting}[style=affectprompt]
Decompose the following two contexts into atomic factual evidence.

Rules:
- One fact per evidence.
- No emotion inference.
- No additional information.
- Generate at most 4 atomic evidence items for Internal Context.
- Generate at most 4 atomic evidence items for External Context.
- If more than 4 facts are available, keep only the 4 most informative and event-relevant facts.
- Use IDs I1, I2, ... for Internal Evidence and E1, E2, ... for External Evidence.

Internal Context:
{internal}

External Context:
{external}

Return JSON:

{
 "internal_evidence":[
  {"id":"I1","text":"..."}
 ],
 "external_evidence":[
  {"id":"E1","text":"..."}
 ]
}
\end{lstlisting}

All internal--external evidence pairs are then scored in one ordered call.

\begin{lstlisting}[style=affectprompt]
Evaluate the support score for EVERY Internal-External evidence pair listed below.

Score meaning:
- 1.0: strongly supporting / describing the same factual clue
- 0.5: compatible or partially supporting
- 0.0: unrelated or conflicting

Use ONLY 0.0, 0.5, or 1.0.
Do NOT use intermediate values.

Pairs are listed in the EXACT output order:
{pair_lines}

There are exactly {n_pairs} pairs.

Output exactly ONE line in this format and nothing else:
SCORES: {score_placeholders}

Requirements:
- After "SCORES:" there MUST be exactly {n_pairs} numeric values separated by commas.
- The k-th value corresponds to the k-th pair above.
- Do not omit any pair.
- Do not output JSON, IDs, explanations, relations, markdown, or any other text.
\end{lstlisting}

\subsection{Event-Conditioned Emotion Inference}
\label{app:prompts_inference}

For each downstream model, we retain the original emotion-classification instruction used by the corresponding baseline and only adapt its output label space to the 11 emotion categories of EGER-Bench. AffectReveal does not replace the baseline-specific classification instruction. Instead, the verified evidence package is appended to the original instruction as auxiliary context, while the original unmasked visual input remains unchanged.

The auxiliary evidence instruction is serialized as follows:

\begin{lstlisting}[style=affectprompt]
External Context:
{external}

Internal Context:
{internal}

Internal-External Evidence Alignment:
Alignment Score: {score}
Aligned Evidence: {aligned_evidence_pairs}
\end{lstlisting}

The resulting auxiliary instruction is concatenated with the original emotion-classification instruction of each downstream model before inference. No additional context-rewriting model is used.

\end{document}